\documentclass{article}

\usepackage[final,nonatbib]{neurips_2019}

\usepackage[utf8]{inputenc} 
\usepackage[T1]{fontenc}    
\usepackage{url}            
\usepackage{amsmath}
\usepackage{graphicx,subcaption}
\usepackage{xcolor}
\usepackage{booktabs}       
\usepackage{amsfonts}       
\usepackage{nicefrac}       
\usepackage{microtype}      
\usepackage{graphicx}
\usepackage{caption,stackengine}

\title{Long-Term Planning and Situational Awareness in OpenAI Five}
\author{
Jonathan Raiman\thanks{equal contribution} \\
Dali\\
\texttt{jonathan@dali.ml} \\
\And
Susan Zhang\footnotemark[1] \\
OpenAI\\
\texttt{susan@openai.com}
\And
Filip Wolski \\
OpenAI\\
\texttt{filip@openai.com}
}

\begin{document}

\maketitle
\begin{abstract}

Understanding how knowledge about the world is represented within model-free deep reinforcement learning methods is a major challenge given the black box nature of its learning process within high-dimensional observation and action spaces. AlphaStar and OpenAI Five have shown that agents can be trained without any explicit hierarchical macro-actions to reach superhuman skill in games that require taking thousands of actions before reaching the final goal.
Assessing the agent's plans and game understanding becomes challenging given the lack of hierarchy or explicit representations of macro-actions in these models, coupled with the incomprehensible nature of the internal representations.

In this paper, we study the distributed representations learned by OpenAI Five to investigate how game knowledge is gradually obtained over the course of training.  We also introduce a general technique for learning a model from the agent's hidden states to identify the formation of plans and subgoals. We show that the agent can learn situational similarity across actions, and find evidence of planning towards accomplishing subgoals minutes before they are executed.  We perform a qualitative analysis of these predictions during the games against the DotA 2 world champions OG in April 2019.
\end{abstract}

\section{Introduction}
The choice of action and plan representation has dramatic consequences on the ability for an agent to explore, learn, or generalize when trying to accomplish a task.  Inspired by how humans methodically organize and plan for long-term goals, Hierarchical Reinforcement Learning (HRL) methods were developed in an effort to augment the set of actions available to the agent to include temporally extended multi-action subroutines.  These extensions help guide exploration and improve sample efficiency, allowing RL agents to tackle problems considered too costly to solve through flat RL \cite{solway2014optimal,botvinick2014model}.

The options framework \cite{sutton1999between} is a popular formulation of HRL that was successfully applied to Atari games in \cite{vezhnevets2017feudal,tomar2018successor}. Kulkarni et al. \cite{kulkarni2016hierarchical} extends upon this framework by not only learning options themselves but also the control policy for composition of options, which allowed for efficient exploration in an environment where feedback was sparse and delayed over long time horizons.
 

Surprisingly, HRL was not involved in training agents to reach superhuman skill at StarCraft and DotA 2, despite games lasting up to an hour long and consisting of thousands of actions. AlphaStar \cite{vinyals2019alphastar} and OpenAI Five \cite{OpenAI_dota} were trained to interact with the games using primitive actions, and were able to overcome the challenges of long-term credit attribution through novel actor-critic algorithms \cite{schulman2017proximal,espeholt2018impala}, large-scale distributed training infrastructure, and self-play.
An important question arises: what internal representations have replaced explicit hierarchy in these models to enable decision-making minutes ahead of time?


To answer this, we examined two aspects of the OpenAI Five model that were not explicitly supervised to contain information about future plans or context. We study whether or not embeddings capture situational similarities, and whether the hidden state of the policy contains information about future milestones or rewards. Our results show that the model does indeed learn semantically meaningful similarities in its distributed vector representation of game attributes, while a new technique for studying hidden states identifies evidence of planning 30-90s ahead of time in OpenAI Five.


\section{Approach}

\subsection{OpenAI Five}

In this paper, we focus specifically on OpenAI Five \cite{OpenAI_dota}, a model trained to reach and surpass professional-level play at the game DotA 2.
DotA 2 is popular multiplayer online battle arena (MOBA) game where two teams of five players play to invade the enemy's base by controlling characters with unique spells and abilities in a fixed map. The game provides a competitive environment for testing planning behavior due to its interest by professional players and 40 minute long average game length.

The OpenAI Five model consists of five LSTMs that each control a particular DotA 2 character in a team, and are trained using a PPO training algorithm \cite{schulman2017proximal} on self-play games against current or past versions of model parameters.
The objective uses a very high discount factor ($
\gamma = 1 - \frac{1}{6300} \approx 0.999841$) to heavily weigh towards future rewards (destroying towers, winning the game) and away from short-term gains (farming, kills). In doing so, the authors were able to train a model to play hour-long games while only back-propagating through 16 timesteps or 2.1 seconds of game time.

\subsection{Similarity Analysis of Distributed Representations}
Even though OpenAI Five was trained with model-free reinforcement learning techniques and without hierarchical representations of plausible action sequences, we noticed that it was capable of transferring its knowledge across changes in the game environment, while also generalizing its learned behaviors against human players. 
We hypothesize that this level of robustness implied the existence of some learned abstraction of fundamental game mechanics, which allowed the agent to adapt to game states that it has never seen before.

Given that the majority of game attributes were fed into the model in the form of embeddings\footnote{for items, actions, heroes, non-heroes, modifiers, abilities, etc.}, it seemed plausible that their corresponding learned distributed representations could also encode functional similarities amongst themselves. Motivated by \cite{wordrep, mikolov2013distributed}, we analyzed the cosine similarity between pairs of embeddings in the final trained OpenAI Five agent to see if an intuitive understanding of their structure could be formed.

\subsection{Hidden State Decoding}

\subsubsection{Future Prediction}


Prior work on evaluating the knowledge of a trained RL agent \cite{jaderberg2016reinforcement,jaderberg2018human} has focused on using linear classifiers on top of the hidden states within the model to produce assessments of the current state of the game. Accuracy of these classifiers can then be used to quantify the agent's understanding of the environment. Unlike those efforts, we would like to predict future game states, along with agent intent towards future courses of action, in order to understand whether or not pre-planning for actions occurred. Because many of these targets are latent or relate to the future, we cannot simply use the current game state as supervision.

\begin{figure}
\centering
\includegraphics[width=\textwidth]{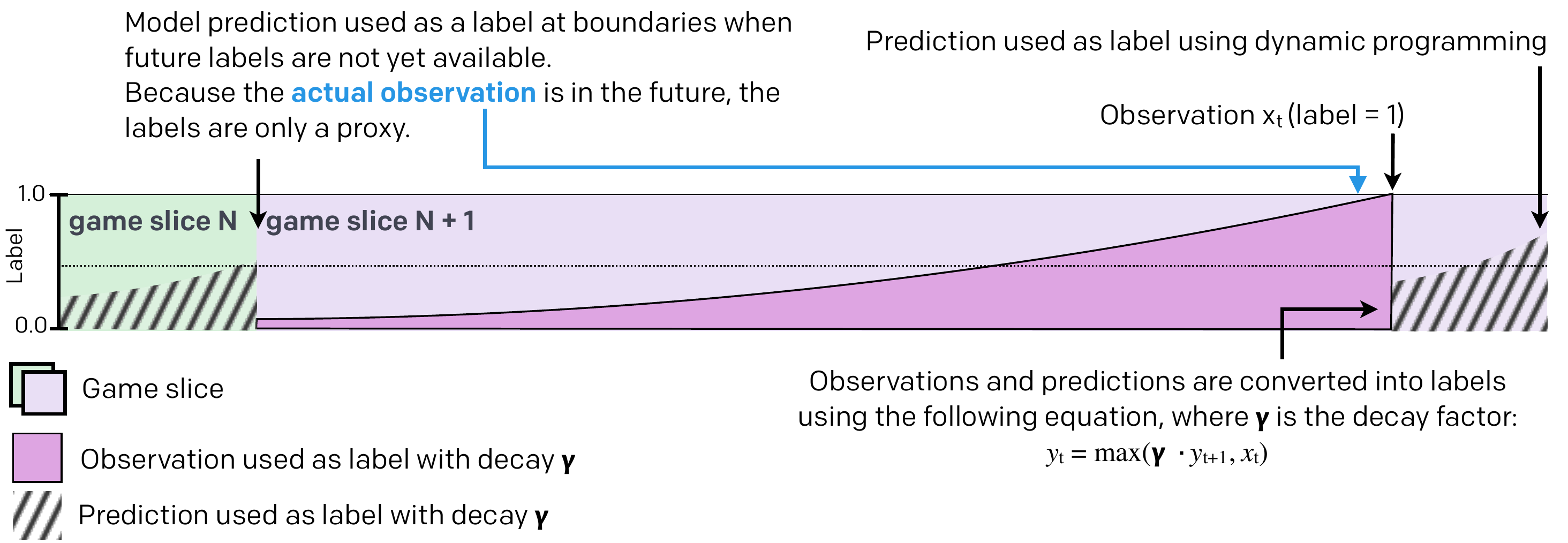}
\caption{\label{fig:supervision}
A learned model makes predictions about future milestones. Because the training of the agent is done over small slices of a full episode, supervision usually requires access to labels that are far in the future. To enable training without seeing future labels, we use dynamic programming, where the present label is a decaying function of the model's prediction at a game slice boundary and the visible observations up to that boundary.
}
\end{figure}

A candidate approach may involve extracting milestones\footnote{e.g. kills, destroying towers, reaching a particular location.} from a collection of saved games and using frozen neural network hidden states from earlier points in the game as input. However, collecting this dataset offline may lead to poor generalization or improperly capture how a model reconsiders its initial plan.  Offline collection is also incompatible with the continuous and asynchronous\footnote{This design choice was done to reduce latency between the parameters in the optimizer and those use in rollouts. Waiting for full games increases the staleness of the rollouts and slows skill progression.} training used in OpenAI Five, since optimization is conducted on 34 second game slices (256 transitions) without waiting for full games to complete, as shown in Figure~\ref{fig:supervision}. By using clipped portions of the game, we cannot obtain online labels about future plans beyond 34 seconds (game time between rollout submissions to the optimizer).

Our solution for online labeling of future plans is to predict discounted sums of future milestones using a decay factor $\gamma$. This allows us to propagate supervision signals across game slices through dynamic programming, similar to what is done for the value function in the Bellman Equation \cite{bellman}. Within a game slice, milestones or rewards are observed as a single scalar and can be discounted starting at their known occurrence location. To supervise predictions about milestones outside of a game slice, we discount the predictions made by the current model at the boundary, as shown in Figure~\ref{fig:supervision}.



\begin{figure}
\centering
\includegraphics[width=0.9\textwidth]{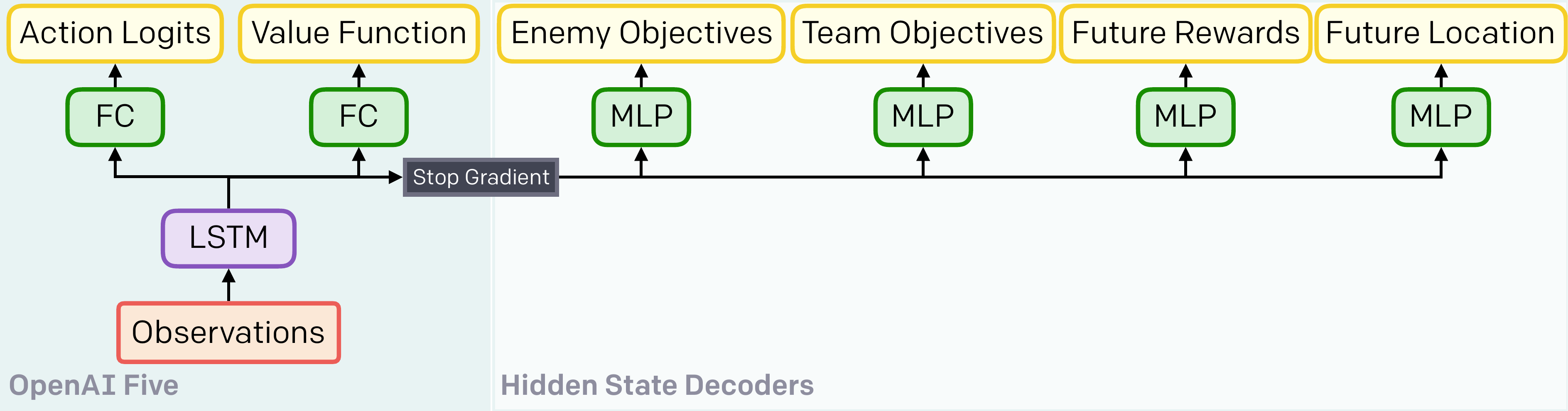}
\caption{\label{fig:arch}
The hidden states from OpenAI Five are passed into Hidden State Decoders that use multi-layer perceptrons (MLP) to predict future observations and milestones. The original model's parameters are supervised by a PPO objective, while the extra predictions are supervised by discounted future milestone labels.
}
\end{figure}

We hypothesize that the LSTM's memory acts as one of the only mechanism for ensuring a consistent set of actions across longer game portions, and thus should contain information about future plans.
To test this hypothesis, we train {\em Hidden State Decoders}: a set of multi-layer perceptrons that receive the LSTM hidden state as input and are taught to predict a continuous value symbolizing how far into the future an event will occur, or the decaying sum of future rewards. Our proposed architecture is illustrated in Figure~\ref{fig:arch}.

We apply this approach to multiple prediction problems within the game. These predictions include rewards (future gold, discounted sum of kill rewards), player's net worth rank (wealth relative to teammates), and milestones (player reaches a map location, team objectives such as when the current player destroys a tower, enemy objectives when the enemy team destroys an allied building).

\subsubsection{Supervision}

We apply slightly different discounting rules when converting milestone and reward observations $x_t$ into labels $y_t$. With milestones, labels $y_t$ at time $t$ are set to 1.0 when the milestone $x_t$ occurs on the next frame, and decayed by some factor $\gamma$ for every frame before: $y_t = \mathrm{max}(\gamma \cdot y_{t+1}, x_t)$ \footnote{The decay factor ensures that our prediction is near 0 when an event is either unlikely to occur or too far into the future, and $\gamma^{T}$ when an event will occur in $T$ steps.}. We then minimize the cross entropy loss between the model's output and the label $y_t$. With reward predictions, we accumulate frame-level observations $x_t$ to create labels using a sum: $y_t = \gamma \cdot y_{t+1} + x_t$. Here, we minimize the mean squared error between model output and $y_t$.


\section{Results}

\subsection{Similarity Analysis of Distributed Representations}

\begin{figure}
\centering
\includegraphics[width=\textwidth]{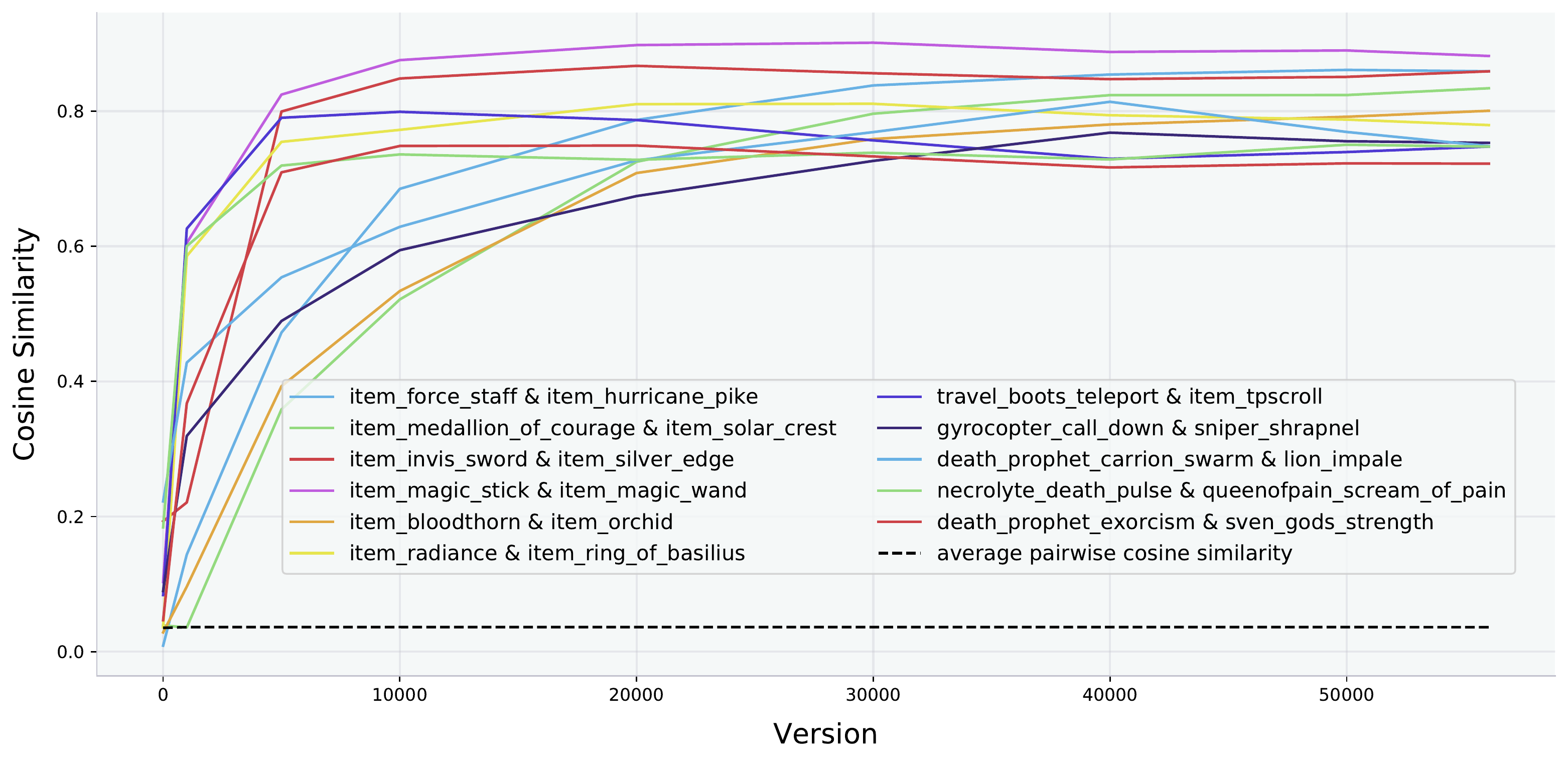}
\caption{\label{fig:cosine}
Evolution of cosine similarity between pairs of actions that are functionally similar within DotA 2. Actions prefixed with \textit{item\_} correspond to the usage of that item within the game.  As the model trained, the cosine similarity measure grew for these functionally similar pairings.
}
\end{figure}

Given a predetermined list of attributes known to be similar within the DotA 2 game, we found that these pairings did indeed exhibit high cosine similarity in their learned vector representations within the final trained OpenAI Five agent.  For these pairings, we also traced their historical similarity values over the course of training.  In Figure~\ref{fig:cosine}, we show a subset of these pairings and their similarity scores.  In the beginning of training (Version 0), all pairings started out equally dissimilar to one another.  Average pairwise similarity lies somewhere around 0.03, which stays mostly constant as the model trained over time. However, after only about 5,000 steps, we can see that the model has quickly learned to distinguish the similarity in these pairings.  Further training pushed these similarity values higher, until they plateau around 30,000 steps.
From these metrics, we see evidence of the model improving its similarity metric for these pairings as training progressed, indicating a growing understanding of how these components relate to one another within the game.

\subsection{Learned Models}

We apply our Hidden State Decoders to different points in training of OpenAI Five. Our predictions discount future milestones and rewards use a decay factor $\gamma = 1 - \frac{1}{900} \approx 0.998$, and we also predict win probability without decay ($\gamma = 1$).

\begin{figure}
\begin{minipage}[t]{0.5\textwidth}
\centering
\includegraphics[height=4cm]{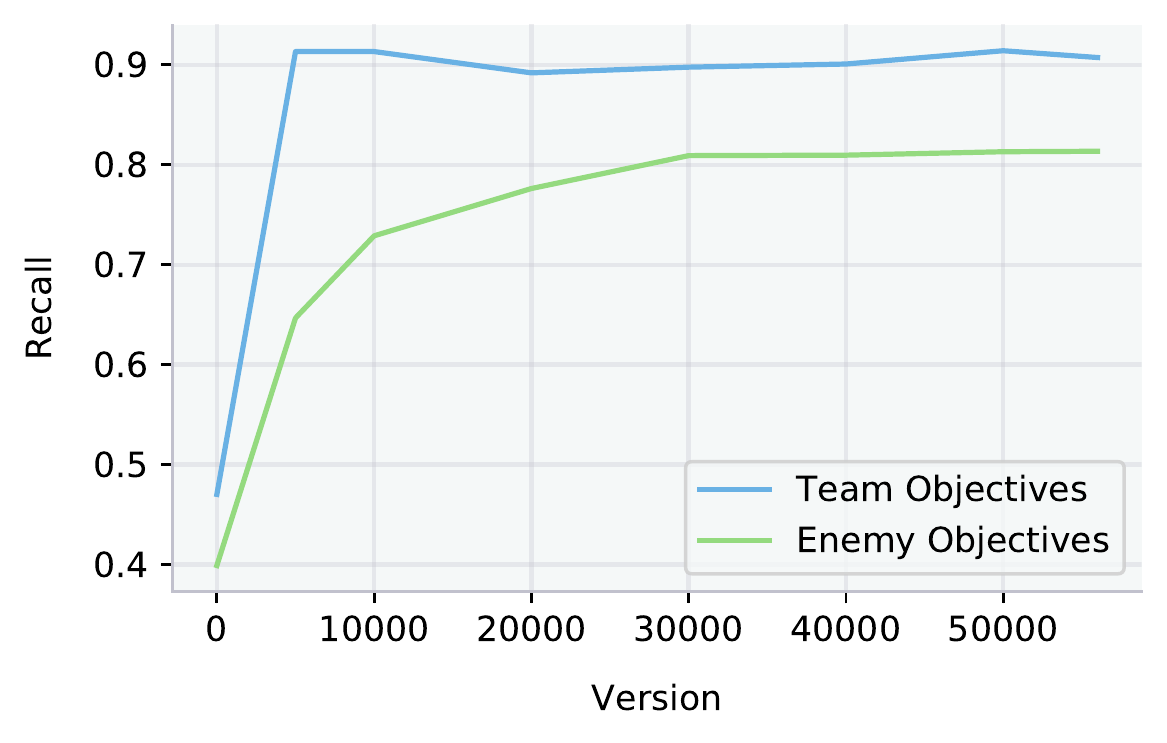}
\end{minipage}
\hfill
\begin{minipage}[t]{0.5\textwidth}
\centering
\includegraphics[height=4cm]{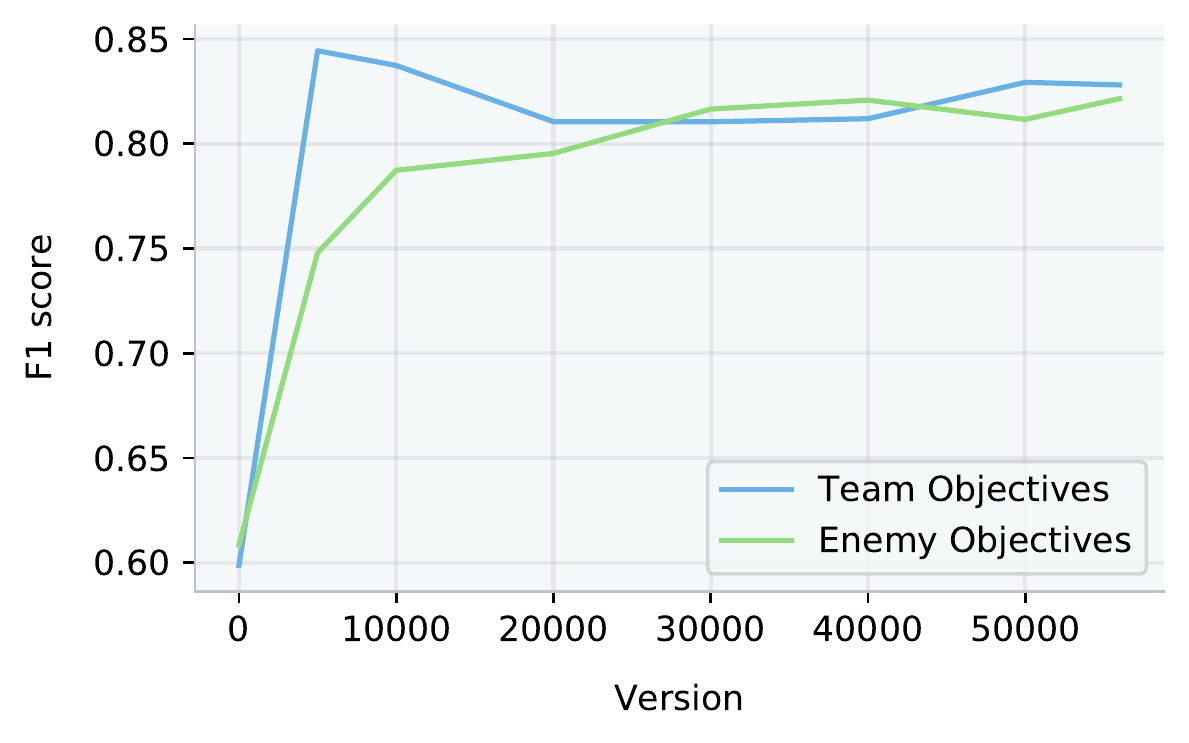}
\end{minipage}
\caption{\label{fig:f1recall}
We compute the F1 score and recall when predicting enemy and team objectives using OpenAI Five hidden states from different points in training. The hidden state gradually contains more information for predicting future objective-taking plans.
}
\end{figure}

After training these hidden state decoders, we compare their predictions on 1,000 self-play games and record the recall and F1 score \footnote{using a threshold selected on a held-out set of 4,000 games} for predicting team or enemy objectives. The learned models reach peak performance for team objective predictions 10,000 versions in, while enemy objective predictions continues to improve in recall and F1 score up to the end of training (56,000 versions in) as visible in Figure~\ref{fig:f1recall}.

\begin{figure}
\centering
\includegraphics[width=0.9\textwidth]{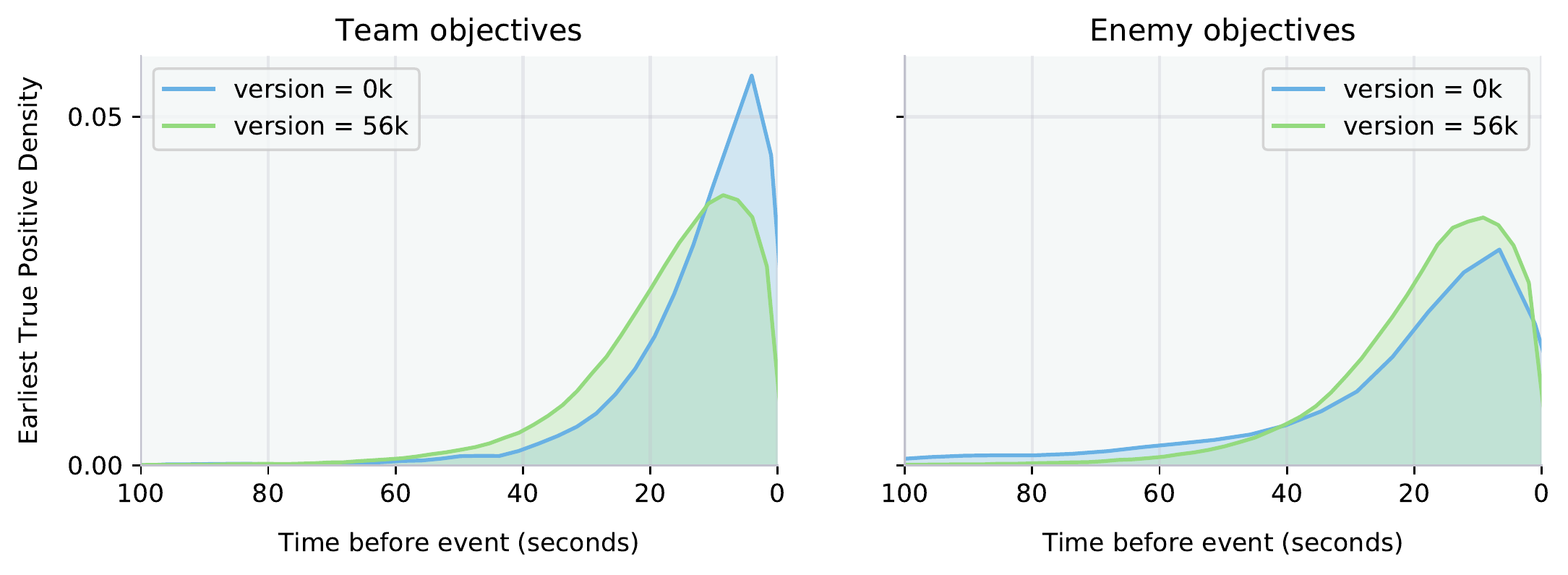}
\caption{\label{fig:earlysign}
We construct a density plot for the earliest point in time the model can correctly predict a future team or enemy objective occurring. With additional training, the model is able to make longer term predictions.
}
\end{figure}

The hidden state decoders produce continuous values corresponding to the discounted milestone observations. We use a threshold to convert these values into a binary classifier for a future milestone to detect how early correct predictions can be made. The density plot of the earliest true positives is shown in Figure~\ref{fig:earlysign}. Note that hidden states from longer trained models better capture information about future plans.

\begin{figure}
\centering
\includegraphics[width=0.8\textwidth]{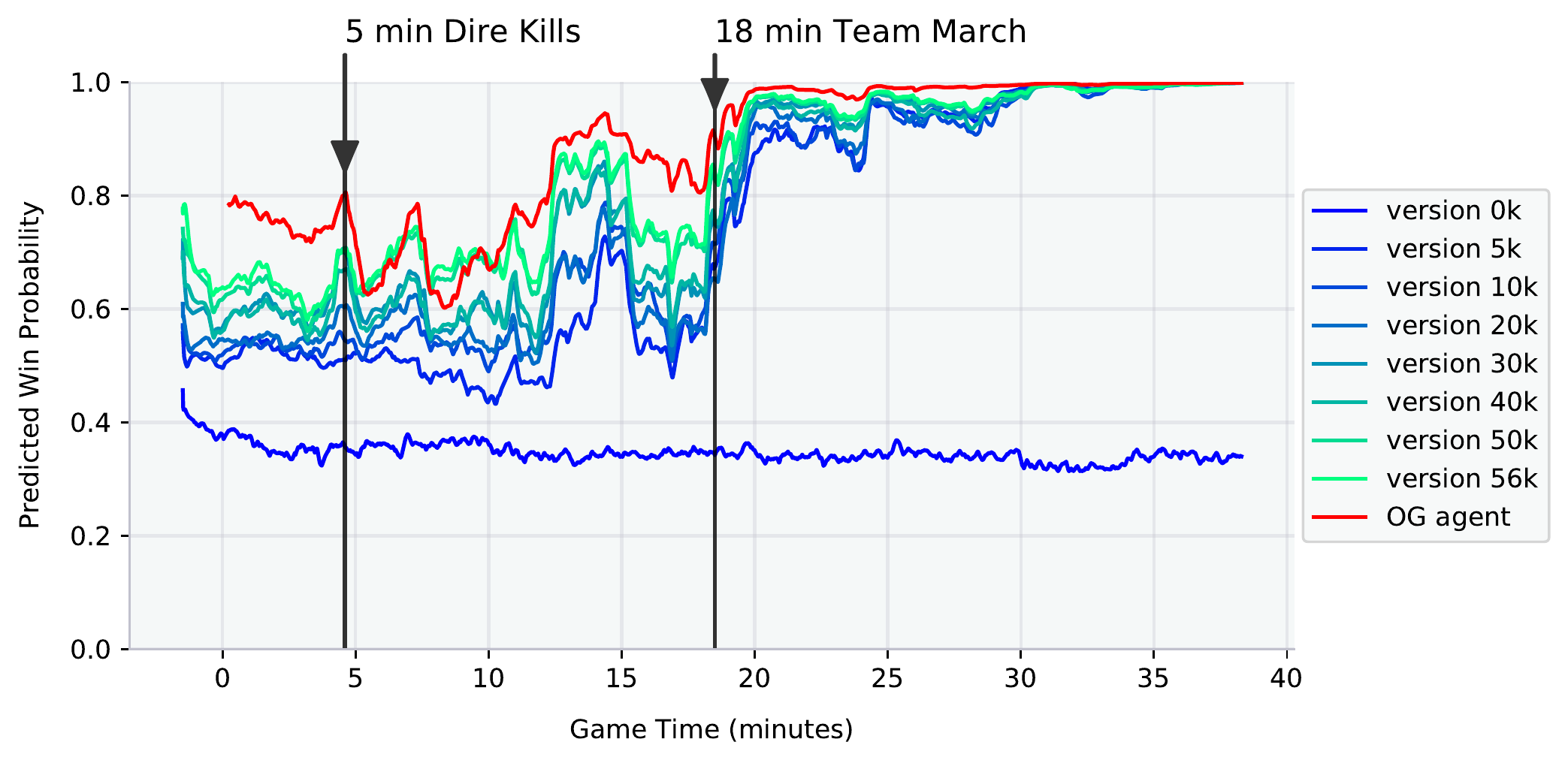}
\caption{
Predicted win probabilities during the first OpenAI Five vs. OG game. The original win probabilities are shown in {\color{red}red}. 
When replaying the games using models trained from scratch up to version 56,000, we see win probabilities gradually approach those given by the model that was used to play OG.
}
\label{fig:winprobs}
\end{figure}

We also compare how OpenAI Five win predictions change across versions when replaying the same game in Figure~\ref{fig:winprobs}. We notice agents earlier in training are less confident in their chances, and as they train, they converge to a similar pattern of confidence as exhibited by the final OpenAI Five model. 

\subsection{Learned Models Interpretation}

In this section we describe how the outputs of the Hidden State Decoders can help interpret and visualize the actions of a trained model-free reinforcement learning agent by observing the predictions during the two OpenAI Five vs. OG games at the Finals event\footnote{Replays and videos with predictions can be downloaded here \url{https://openai.com/blog/how-to-train-your-openai-five/#replays}.}.

\begin{figure}
\begin{subfigure}[t]{\linewidth}
\centering
\includegraphics[width=\textwidth]{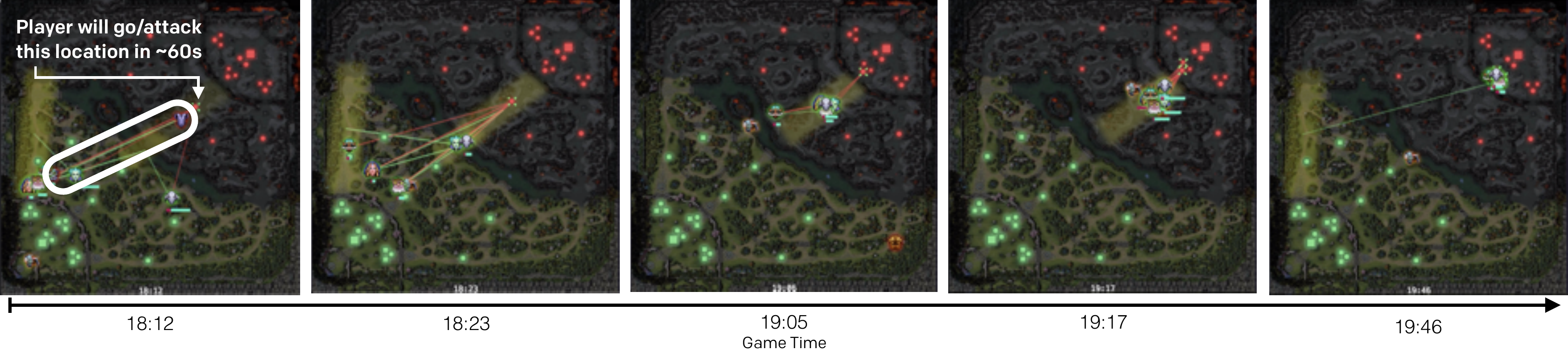}
\caption{\label{fig:push}}
\end{subfigure}
\begin{subfigure}[t]{\linewidth}
\centering
\includegraphics[width=0.6\textwidth]{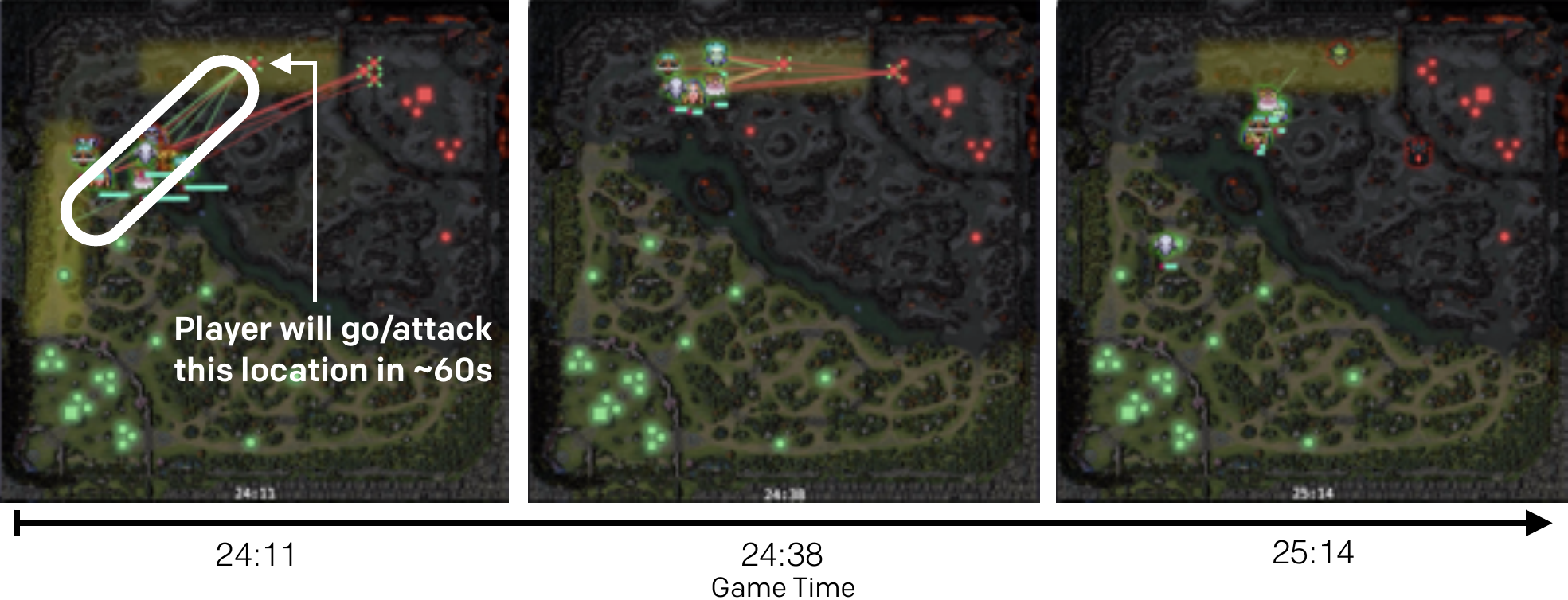}
\caption{\label{fig:bail}}
\end{subfigure}
\caption{
Objective Predictions: a red line is drawn from player to enemy buildings that will be attacked in $\leq60$ seconds. Green lines points to map regions the player will reach in $\leq60$ seconds.\\
{\bf Game 1 (\ref{fig:push})}: The model predicts the Radiant (bottom) team will attack the Dire (top) towers in the middle of the map. 55 seconds later, the team is about to accomplish this. At this point, the predictions also include destroying towers within the enemy base, and 29s later, the team attacks those towers.\\
{\bf Game 2 (\ref{fig:bail})}: The model predicts the Radiant (bottom) team will attack a tower at the top of the map. 27s later, the team has approached the tower, and the prediction now includes attacking a tower on the left side of the enemy base. 36s later, an enemy hero comes to defend this tower, and the model stops predicting a continued attack.
}
\end{figure}

We can visually connect the location of each character on a map of the game to the location they expect to reach or enemy building they will attack in the near future. When this information is displayed for the entire team, it is possible to detect early signs of an attack and its target. We demonstrate this in Figure~\ref{fig:push}. The model predictions are sometimes incorrect due to partial information and unpredictability of the enemy players. We can see evidence of the predictions changing dramatically based on the enemy team defending their position in Figure~\ref{fig:bail}.

\begin{figure}
\centering
\includegraphics[width=\textwidth]{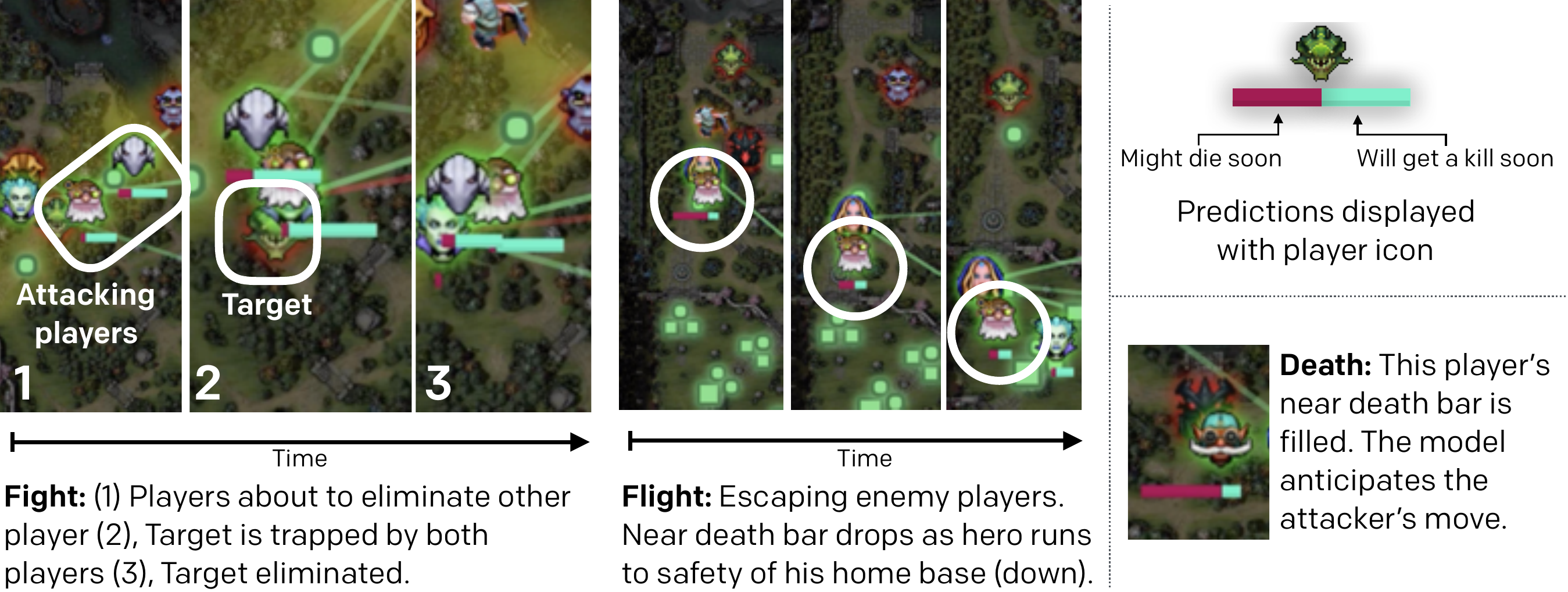}
\caption{\label{fig:fight}
Model predicts kills and deaths. The near death prediction spikes during ``flight'' behavior, while the kill prediction spikes when one or more players find a target to attack. Note however that the predictions are anticipatory: in the {\bf Flight} images the character narrowly escapes his attackers, despite initially predicting it will die.
}
\end{figure}

Death and kill reward prediction also helps to comprehend the reasoning behind model actions. We train our hidden state decoders to predict future rewards from kills and penalties from dying as shown in Figure~\ref{fig:fight}. The predictions are near zero when death is unlikely or too far in the future, and fire when there is a high chance of dying or killing soon.

Combined with movement prediction, we can use the death and kill information to know whether a particular action is used offensively, defensively, or passively. In Figure~\ref{fig:fight} the {\bf Flight} series of screenshots show a character escaping enemy players while the death prediction decreases, indicating an evasive maneuver. We see the opposite pattern emerge in {\bf Fight} where the kill prediction gauge grows as two characters close in on an enemy player.

\section{Conclusion}

OpenAI Five has demonstrated, contrary to prior belief, the emergence of macro-action representations and goals, as well as preliminary evidence of a semantic understanding of game attributes, despite being trained to maximize reward without any hierarchical organization of the action space. We introduce two new methodologies for extracting the knowledge representation discovered by the model: (1) evolution of cosine similarities between embedding vectors over the course of training, and (2) a supervision and training methodology for decoding model hidden states into near and long-term goals. Through analysis of the model parameters over the course of training, along with the games against OG on April 13th, we have shown empirical evidence of planning future objectives 30-90 seconds ahead of time, and an ability to rationalise evasive and offensive maneuvers through a fight or flight predictor. 
We believe the presented techniques can generalize to other large scale RL agents that operate in a semantically rich environment, extracting knowledge about plans in situations where hundreds to thousands of actions are required to accomplish a task.

As future work, we believe the methods introduced in this paper could be used to augment collaboration between artificial intelligence agents and humans by creating a shared interface for control. Early signs of this direction were demonstrated during the cooperative game at the OpenAI Finals event\footnote{Game recording available here \url{https://openai.com/blog/how-to-train-your-openai-five/#cooperativemode}}, where human players played alongside trained agents on the same team.  In these games, the trained agents were able to communicate aspects of their plans to their fellow human team members, which enabled potential collaboration between the two groups.

\bibliographystyle{unsrt}
\bibliography{bibliography}

\begin{thebibliography}{10}

\bibitem{solway2014optimal}
Alec Solway, Carlos Diuk, Natalia C{\'o}rdova, Debbie Yee, Andrew~G Barto, Yael
  Niv, and Matthew~M Botvinick.
\newblock Optimal behavioral hierarchy.
\newblock {\em PLoS computational biology}, 10(8):e1003779, 2014.

\bibitem{botvinick2014model}
Matthew Botvinick and Ari Weinstein.
\newblock Model-based hierarchical reinforcement learning and human action
  control.
\newblock {\em Philosophical Transactions of the Royal Society B: Biological
  Sciences}, 369(1655):20130480, 2014.

\bibitem{sutton1999between}
Richard~S Sutton, Doina Precup, and Satinder Singh.
\newblock Between mdps and semi-mdps: A framework for temporal abstraction in
  reinforcement learning.
\newblock {\em Artificial intelligence}, 112(1-2):181--211, 1999.

\bibitem{vezhnevets2017feudal}
Alexander~Sasha Vezhnevets, Simon Osindero, Tom Schaul, Nicolas Heess, Max
  Jaderberg, David Silver, and Koray Kavukcuoglu.
\newblock Feudal networks for hierarchical reinforcement learning.
\newblock In {\em Proceedings of the 34th International Conference on Machine
  Learning-Volume 70}, pages 3540--3549. JMLR. org, 2017.

\bibitem{tomar2018successor}
Manan Tomar, Rahul Ramesh, and Balaraman Ravindran.
\newblock Successor options: An option discovery algorithm for reinforcement
  learning.
\newblock In {\em Proceedings of the Twenty-Eighth International Joint
  Conference on Artificial Intelligence (IJCAI-19)}, 2019.

\bibitem{kulkarni2016hierarchical}
Tejas~D Kulkarni, Karthik Narasimhan, Ardavan Saeedi, and Josh Tenenbaum.
\newblock Hierarchical deep reinforcement learning: Integrating temporal
  abstraction and intrinsic motivation.
\newblock In {\em Advances in neural information processing systems}, pages
  3675--3683, 2016.

\bibitem{vinyals2019alphastar}
Oriol Vinyals, Igor Babuschkin, Junyoung Chung, Michael Mathieu, Max Jaderberg,
  Wojciech~M Czarnecki, Andrew Dudzik, Aja Huang, Petko Georgiev, Richard
  Powell, et~al.
\newblock Alphastar: Mastering the real-time strategy game starcraft ii.
\newblock {\em DeepMind Blog}, 2019.

\bibitem{OpenAI_dota}
OpenAI.
\newblock Openai five.
\newblock \url{https://blog.openai.com/openai-five/}, 2018.

\bibitem{schulman2017proximal}
John Schulman, Filip Wolski, Prafulla Dhariwal, Alec Radford, and Oleg Klimov.
\newblock Proximal policy optimization algorithms.
\newblock {\em arXiv preprint arXiv:1707.06347}, 2017.

\bibitem{espeholt2018impala}
Lasse Espeholt, Hubert Soyer, Remi Munos, Karen Simonyan, Volodymir Mnih, Tom
  Ward, Yotam Doron, Vlad Firoiu, Tim Harley, Iain Dunning, et~al.
\newblock Impala: Scalable distributed deep-rl with importance weighted
  actor-learner architectures.
\newblock {\em arXiv preprint arXiv:1802.01561}, 2018.

\bibitem{wordrep}
Tomas Mikolov, Kai Chen, Greg Corrado, and Jeffrey Dean.
\newblock Efficient estimation of word representations in vector space.
\newblock {\em ICLR Workshop}, 2013.

\bibitem{mikolov2013distributed}
Tomas Mikolov, Ilya Sutskever, Kai Chen, Greg~S Corrado, and Jeff Dean.
\newblock Distributed representations of words and phrases and their
  compositionality.
\newblock In {\em Advances in neural information processing systems}, pages
  3111--3119, 2013.

\bibitem{jaderberg2016reinforcement}
Max Jaderberg, Volodymyr Mnih, Wojciech~Marian Czarnecki, Tom Schaul, Joel~Z
  Leibo, David Silver, and Koray Kavukcuoglu.
\newblock Reinforcement learning with unsupervised auxiliary tasks.
\newblock {\em arXiv preprint arXiv:1611.05397}, 2016.

\bibitem{jaderberg2018human}
Max Jaderberg, Wojciech~M Czarnecki, Iain Dunning, Luke Marris, Guy Lever,
  Antonio~Garcia Castaneda, Charles Beattie, Neil~C Rabinowitz, Ari~S Morcos,
  Avraham Ruderman, et~al.
\newblock Human-level performance in first-person multiplayer games with
  population-based deep reinforcement learning.
\newblock {\em arXiv preprint arXiv:1807.01281}, 2018.

\bibitem{bellman}
Richard~E. Bellman.
\newblock {\em Dynamic Programming}.
\newblock Princeton University Press, Princeton, NJ, 1957.

\end{thebibliography}

\end{document}